\documentclass[12pt]{iopart}
\usepackage{amssymb}
\usepackage{adjustbox}
\usepackage{array}
\usepackage{xcolor}
\usepackage{lmodern}
\usepackage{booktabs}
\usepackage{longtable}
\usepackage{colortbl}
\usepackage{pdflscape} 
\usepackage{rotating} 
\usepackage[a4paper, margin=1in]{geometry}
\usepackage{afterpage}
\usepackage{lscape}

\usepackage{natbib}
  \expandafter\let\csname equation*\endcsname\relax
  \expandafter\let\csname endequation*\endcsname\relax
\usepackage{graphics}
\usepackage{graphicx}
\usepackage{caption}
\usepackage{color}
\usepackage{soul}
\usepackage{subfig}
\usepackage{tabularx}
\usepackage{multirow}
\setstcolor{red}
\bibpunct{[}{]}{,}{n}{,}{,}
\usepackage{svg}
\usepackage{hyperref}
\hypersetup{
    colorlinks=true,
    linkcolor=blue,
    filecolor=magenta,      
    urlcolor=blue,
    citecolor=blue 
}
\begin{document}

\title{Decade of Natural Language Processing in Chronic Pain: A Systematic Review}
\label{title}

\author{ 
Swati Rajwal$^{1}$
\\
\small{
$^1$ Department of Biomedical Informatics, Emory University, USA\\
}}

\ead{swati.rajwal@emory.edu}

\begin{abstract}
\label{abstract}
In recent years, the intersection of Natural Language Processing (NLP) and public health has opened innovative pathways for investigating various domains, including chronic pain in textual datasets. Despite the promise of NLP in chronic pain, the literature is dispersed across various disciplines, and there is a need to consolidate existing knowledge, identify knowledge gaps in the literature, and inform future research directions in this emerging field. This review aims to investigate the state of the research on NLP-based interventions designed for chronic pain research. A search strategy was formulated and executed across PubMed, Web of Science, IEEE Xplore, Scopus, and ACL Anthology to find studies published in English between 2014 and 2024. After screening 132 papers, 26 studies were included in the final review. Key findings from this review underscore the significant potential of NLP techniques to address pressing challenges in chronic pain research. The past 10 years in this field have showcased the utilization of advanced methods (transformers like RoBERTa and BERT) achieving high-performance metrics (e.g., F\textsubscript{1}$>$0.8) in classification tasks, while unsupervised approaches like Latent Dirichlet Allocation (LDA) and k-means clustering have proven effective for exploratory analyses. Results also reveal persistent challenges such as limited dataset diversity, inadequate sample sizes, and insufficient representation of underrepresented populations. Future research studies should explore multimodal data validation systems, context-aware mechanistic modeling, and the development of standardized evaluation metrics to enhance reproducibility and equity in chronic pain research.
\\


\end{abstract}


\section{Introduction}
\subsection{Rationale}
\label{rationale}
Chronic pain refers to the condition where pain exists for more than 3 months on most days \cite{rikard_chronic_2023}. Approximately 12 million U.S. adults suffer from both chronic pain and significant anxiety or depression. Over half of those with chronic pain also report persistent mental health symptoms, and nearly 70\% say health issues limit their work, daily tasks, and social activities \cite{de_la_rosa_co-occurrence_2024}. Studies have shown that innovative technologies like Natural Language Processing (NLP) are emerging as promising tools in the field of chronic pain research and treatment \cite{bacco_natural_2022}. NLP is a branch of artificial intelligence that focuses on the interaction between computers and human language. NLP has shown potential in various aspects of chronic pain management. Recent studies have demonstrated its effectiveness in analyzing patient-reported outcomes, identifying imaging findings related to low back pain, and even predicting placebo analgesia in chronic pain patients \cite{bacco_natural_2022,branco_predicting_2023}.

\subsection{Objectives}
\label{objectives}
Despite the promise of NLP in chronic pain, the literature is dispersed, and therefore, a systematic review is necessary to consolidate existing knowledge, identify knowledge gaps in the literature, and inform future research directions in this emerging field. This systematic review has two major objectives. First, to identify and highlight distinct NLP techniques (including Large Language Models) used for tasks related to Chronic pain. The second objective is to report the effectiveness of such techniques/models, identify potential knowledge gaps, and design research questions for future studies.

\section{Methods}
This systematic review is carried out under the PRISMA (Preferred
Reporting Items for Systematic Reviews and Meta-analysis)
guidelines \cite{page_prisma_2021} and the checklist can be found in \ref{checklist_table}.

\subsection{Eligibility Criteria}
\label{eligibility_criteria}
The eligible publications for this review are restricted to peer-reviewed published literature (observational studies, algorithm validation studies, computational model evaluations, experimental, qualitative), including journal articles and full conference papers. The study must be written in English, although the language of the textual dataset used can vary. The publication period of the included studies was restricted to the last decade, i.e., from 2014 to 2024. To be included, a study should answer a research question(s) on the design, development, and application of NLP in chronic pain.



\subsection{Information Sources}
\label{information_sources}
A systematic search of the following databases was conducted from 01\textsuperscript{st} January 2014 until 15\textsuperscript{th} September 2024 date across PubMed, Web of Science, IEEE Xplore, Scopus, and ACL Anthology. The selected databases offer comprehensive coverage of both healthcare and NLP research. PubMed and Web of Science provide robust access to biomedical literature on chronic pain, while IEEE Xplore and ACL Anthology focus on technical advancements in NLP. Scopus bridges these disciplines, ensuring an interdisciplinary approach for a thorough systematic review. Preprints (arxiv/biorxiv), forewords, prefaces, table of contents, programs, schedules, indexes, call for papers/participation, lists of reviewers, lists of tutorial abstracts, invited talks, appendices, session information, obituaries, book reviews, newsletters, lists of proceedings, lifetime achievement awards, erratum, systematic reviews, scoping reviews, and notes are excluded. Finally, backward and forward citation chasing was performed on all the selected studies using an R-package called \textit{Citationchaser} as introduced in the study by Haddaway \& colleagues \cite{noauthor_citationchaser_nodate}.

\subsection{Search Strategy}
\label{search_strategy}
Three keywords were used ``chronic pain", ``natural language processing" and ``large language model". The search results across databases are shown in Table~\ref{search_results_tab}. The full search strategies for all information sources are provided in~\ref{search_strategy_app}.

\subsection{Selection Process}
\label{selection_process}
One reviewer (SR) independently screened each study for eligibility by marking it as a ‘yes’ (for inclusion), ‘no’ (for exclusion), or ‘maybe’ (in case of uncertainty about relevance) in the Covidence platform (https://app.covidence.org). Full access to the Covidence platform was provided via Emory University login. In the first stage, the reviewer screened the titles and abstracts of each study as identified in the databases by the search strategies. In the second stage of screening, the full-text manuscripts were screened as per the eligibility criteria. Studies that do not meet the eligibility criteria were moved to an exclusion folder.

\subsection{Data Collection Process}
\label{data_collection_process}
One independent researcher (SR) extracted data from the final included full-texts. Before formal data extraction, the data extraction form was piloted with a sample paper to identify and address any issues in the form to ensure it is comprehensive. The data extraction was then conducted on all papers.

\subsection{Data Items}
\label{data_items}
For each selected article, data items were extracted, such as year of publication, study design, research question, dataset description, NLP technology used, number of participants and their age range, results, and reproducibility. 

\subsection{Study risk of bias assessment}
\label{study_bias_method}
The risk of bias in the included studies was assessed using a simplified checklist inspired by the Joanna Briggs Institute (JBI) Critical Appraisal Tools \cite{noauthor_jbi_nodate}. The checklist was tailored to evaluate studies employing NLP techniques for tasks related to chronic pain. The checklist included 11 items across four domains: (1) study objectives and context, (2) study design and data, (3) model development and evaluation, and (4) results and interpretation. For each item, studies were rated as ``Yes," ``No", or ``Unclear." A composite score was then calculated to classify the overall risk of bias into three categories: low ($\geq 8$ `Yes'), moderate (5-7 `Yes'), or high ($\leq 5$ `Yes'). 

\subsection{Reporting bias assessment}
\label{bias_assessment}
To assess the risk of bias due to missing results in the synthesis, a customized checklist was developed, inspired by the ROBINS-I tool \cite{sterne_robins-i_2016}. This checklist was modified to evaluate reporting bias in studies employing NLP techniques for chronic pain and focused on the completeness of results reporting selective reporting, and transparency. Each study was assessed using this checklist, with responses recorded as ``Yes," ``No," or ``Unclear." A composite score was calculated for each study to classify the risk of reporting bias as low risk (7–8 ``Yes"), moderate risk (4–6 ``Yes"), and high risk ($<$4 ``Yes"). No formal statistical methods were applied to detect publication bias due to the heterogeneity of the included studies. 

\subsection{Study Records}
\label{study_records}
The search query results from each database were exported in RIS format (except for ACL anthology since it has no way to batch export records) and then imported into Covidence Software which removed all the duplicates.

\subsection{Ethical Considerations}
This review involved the analysis of previously published work and did not require ethical approval or patient consent. Study participants’ confidentiality
and anonymity was maintained by reporting aggregated data without individual identifiers - per the included studies.

\section{Results}
\begin{figure*}
  \centering \includegraphics[width=1.0\textwidth]{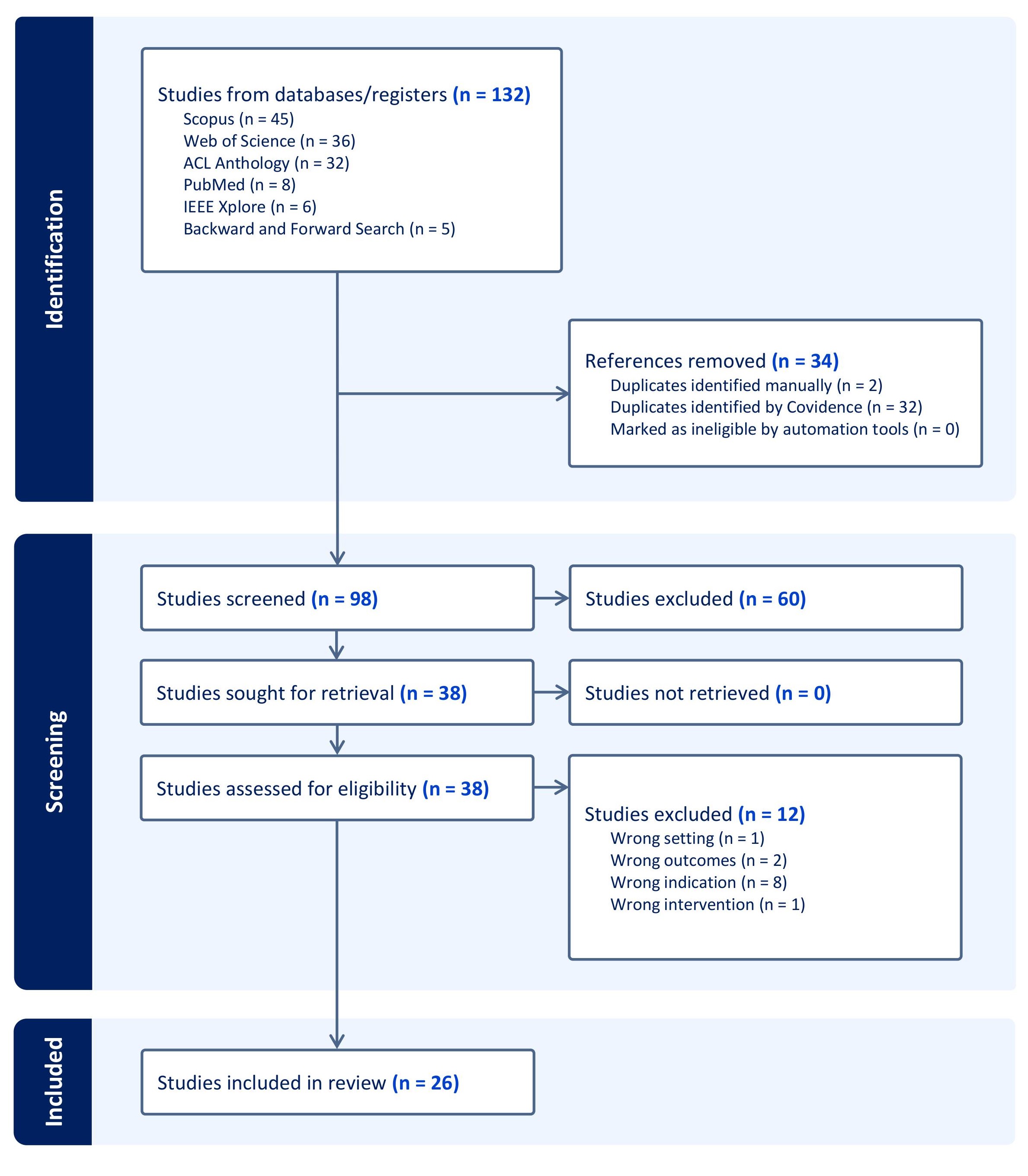}
  \caption{PRISMA flow diagram.}
  \label{fig_prisma}
\end{figure*}

\subsection{Study Selection}
\label{study_selection}
A total of 127 studies were initially identified through database searching, which included 45 from Scopus, 36 from Web of Science, 32 from ACL Anthology, 8 from PubMed, and 6 from IEEE Xplore. After doing one round of eligibility run, 5 additional studies were added through backward and forward search, showing 132 studies in total. No additional references were identified through citation searching or grey literature sources. After removing 30 duplicates, identified automatically using Covidence, 97 studies remained for screening. During the screening phase, 60 studies were excluded based on the inclusion and exclusion criteria. A total of 38 studies were sought for full-text retrieval, and all were successfully retrieved for further eligibility assessment. Of these 38 studies, 12 were excluded for the following reasons: 1 for wrong setting, 2 for wrong outcomes, 8 for wrong indication, and 1 for wrong intervention. Finally, 26 studies were included in the final review. No ongoing studies or studies awaiting classification were identified at the time of this review. The detailed process can be found in the PRISMA flow diagram in Figure~\ref{fig_prisma}.

\subsection{Study Characteristics}
\label{study_char}
\begin{figure*}
  \centering \includegraphics[width=0.8\textwidth]{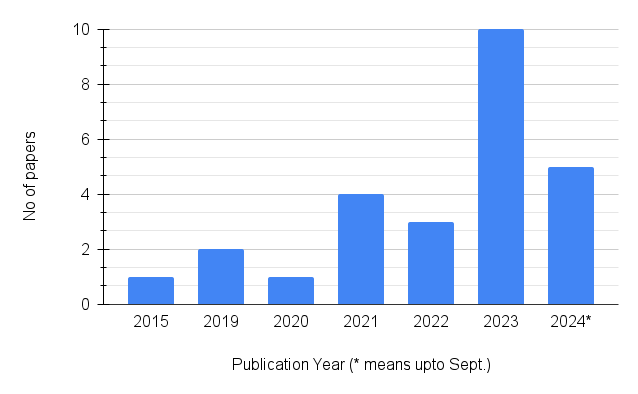}
  \caption{Number of articles per year considered in this review.}
  \label{review_studies_freq}
\end{figure*}

A total of 26 studies were selected after thoroughly screening titles, abstracts, and full texts. These studies were published between 2015 and 2024 (Figure~\ref{review_studies_freq}).
The studies were primarily published in the past three years, highlighting the growing interest in applying NLP techniques to chronic pain. The research spans various topics, including predicting placebo analgesia, developing annotated corpora for pain-related language, identifying language features in placebo studies, and modeling chronic pain experiences using online narratives.

\subsection{Risk of bias in studies}
\label{risk_of_bias}
\begin{table}
\centering
\caption{Risk of Bias Assessment for Included Studies.\newline{\color{green} \checkmark} = `Yes', {\color{red} $\times$} = `No', {\color{blue} $\bullet$} = `Unclear', {\textbf{$\downarrow$}} = `Low risk of bias', {\textbf{$\approx$}} = `Moderate risk of bias'.}
\label{risk_bias_table}
\renewcommand{\arraystretch}{1.8} 
\resizebox{\textwidth}{!}{%
\fontsize{25}{25}\selectfont 
\begin{tabular}{p{8cm}llllllllllllllllllllllllll}
\toprule
 & \Large{\cite{necaise_peer_2024}} & \Large{\cite{lotz_exploration_2023}} & \Large{\cite{nunes_modeling_2023}} & \Large{\cite{chaturvedi_distributions_2024}} & \Large{\cite{venerito_large_2024}} & \Large{\cite{j_m_reinen_remotely-captured_2024}} & \Large{\cite{sarker_chronicpain_2023}} & \Large{\cite{himmelstein_examination_2022}} & \Large{\cite{c_agurto_exploring_2024}} & \Large{\cite{nunes_chronic_2023}} & \Large{\cite{aggarwal_cross-modal_2023}} & \Large{\cite{gordon_relationship_2023}} & \Large{\cite{ashar_yk_reattribution_2023}} & \Large{\cite{dobscha_mental_2023}} & \Large{\cite{goldstein_characterizing_2023}} & \Large{\cite{schirle_two_2021}} & \Large{\cite{chaturvedi_development_2023}} & \Large{\cite{chen_understanding_2019}} & \Large{\cite{hylan_automated_2015}} & \Large{\cite{berger_quantitative_2021}} & \Large{\cite{taylor_use_2019}} & \Large{\cite{chaturvedi_development_2021}} & \Large{\cite{davidson_job-related_2021}} & \Large{\cite{branco_predicting_2022}} & \Large{\cite{goudman_social_2022}} & \Large{\cite{yang_combining_2020}} \\ 
\midrule
\Large{Is study aim clearly stated?} & {\color{green} \checkmark} &	{\color{green} \checkmark} &	{\color{green} \checkmark} &	{\color{green} \checkmark} &	{\color{green} \checkmark} &	{\color{green} \checkmark} &	{\color{green} \checkmark} &	{\color{green} \checkmark} &	{\color{green} \checkmark} &	{\color{green} \checkmark} &	{\color{green} \checkmark} &	{\color{green} \checkmark} &	{\color{green} \checkmark} &	{\color{green} \checkmark} &	{\color{green} \checkmark} &	{\color{green} \checkmark} &	{\color{green} \checkmark} &	{\color{green} \checkmark} &	{\color{green} \checkmark} &	{\color{green} \checkmark} &	{\color{green} \checkmark} &	{\color{green} \checkmark} &	{\color{green} \checkmark} &	{\color{green} \checkmark} &	{\color{green} \checkmark} &	{\color{green} \checkmark} \\ 
\midrule
\Large{Is the study relevant to chronic pain and NLP?} & {\color{green} \checkmark} &	{\color{green} \checkmark} &	{\color{green} \checkmark} &	{\color{green} \checkmark} &	{\color{green} \checkmark} &	{\color{green} \checkmark} &	{\color{green} \checkmark} &	{\color{green} \checkmark} &	{\color{green} \checkmark} &	{\color{green} \checkmark} &	{\color{green} \checkmark} &	{\color{green} \checkmark} &	{\color{green} \checkmark} &	{\color{green} \checkmark} &	{\color{green} \checkmark} &	{\color{green} \checkmark} &	{\color{green} \checkmark} &	{\color{green} \checkmark} &	{\color{green} \checkmark} &	{\color{green} \checkmark} &	{\color{green} \checkmark} &	{\color{green} \checkmark} &	{\color{green} \checkmark} &	{\color{green} \checkmark} &	{\color{green} \checkmark} &	{\color{green} \checkmark}  \\ \midrule
\Large{Is study design appropriate for evaluating the NLP technique?} & {\color{green} \checkmark} &	{\color{blue} $\bullet$} &	{\color{green} \checkmark} &	{\color{green} \checkmark} &	{\color{blue} $\bullet$} &	{\color{green} \checkmark} &	{\color{green} \checkmark} &	{\color{green} \checkmark} &	{\color{green} \checkmark} &	{\color{blue} $\bullet$} &	{\color{green} \checkmark} &	{\color{blue} $\bullet$} &	{\color{green} \checkmark} &	{\color{green} \checkmark} &	{\color{green} \checkmark} &	{\color{green} \checkmark} &	{\color{green} \checkmark} &	{\color{green} \checkmark} &	{\color{green} \checkmark} &	{\color{green} \checkmark} &	{\color{green} \checkmark} &	{\color{red} $\times$} &	{\color{green} \checkmark} &	{\color{green} \checkmark} &	{\color{green} \checkmark} &	{\color{green} \checkmark}  \\ \midrule
\Large{Are the data sources described adequately?} & {\color{green} \checkmark} &	{\color{blue} $\bullet$} &	{\color{green} \checkmark} &	{\color{green} \checkmark} &	{\color{green} \checkmark} &	{\color{green} \checkmark} &	{\color{green} \checkmark} &	{\color{green} \checkmark} &	{\color{green} \checkmark} &	{\color{green} \checkmark} &	{\color{green} \checkmark} &	{\color{green} \checkmark} &	{\color{green} \checkmark} &	{\color{green} \checkmark} &	{\color{green} \checkmark} &	{\color{green} \checkmark} &	{\color{green} \checkmark} &	{\color{green} \checkmark} &	{\color{green} \checkmark} &	{\color{green} \checkmark} &	{\color{green} \checkmark} &	{\color{green} \checkmark} &	{\color{green} \checkmark} &	{\color{green} \checkmark} &	{\color{green} \checkmark} &	{\color{green} \checkmark} \\ \midrule
\Large{Is the NLP technique adequately described?} & {\color{green} \checkmark} &	{\color{green} \checkmark} &	{\color{green} \checkmark} &	{\color{blue} $\bullet$} &	{\color{green} \checkmark} &	{\color{blue} $\bullet$} &	{\color{blue} $\bullet$} &	{\color{blue} $\bullet$} &	{\color{blue} $\bullet$} &	{\color{green} \checkmark} &	{\color{green} \checkmark} &	{\color{blue} $\bullet$} &	{\color{blue} $\bullet$} &	{\color{blue} $\bullet$} &	{\color{blue} $\bullet$} &	{\color{green} \checkmark} &	{\color{blue} $\bullet$} &	{\color{green} \checkmark} &	{\color{blue} $\bullet$} &	{\color{green} \checkmark} &	{\color{green} \checkmark} &	{\color{blue} $\bullet$} &	{\color{green} \checkmark} &	{\color{green} \checkmark} &	{\color{green} \checkmark} &	{\color{green} \checkmark} \\\midrule

\Large{Are evaluation metrics clearly reported?} & {\color{green} \checkmark} &	{\color{green} \checkmark} &	{\color{green} \checkmark} &	{\color{green} \checkmark} &	{\color{green} \checkmark} &	{\color{blue} $\bullet$} &	{\color{green} \checkmark} &	{\color{blue} $\bullet$} &	{\color{green} \checkmark} &	{\color{green} \checkmark} &	{\color{green} \checkmark} &	{\color{green} \checkmark} &	{\color{green} \checkmark} &	{\color{green} \checkmark} &	{\color{green} \checkmark} &	{\color{green} \checkmark} &	{\color{green} \checkmark} &	{\color{green} \checkmark} &	{\color{green} \checkmark} &	{\color{green} \checkmark} &	{\color{green} \checkmark} &	{\color{red} $\times$} &	{\color{red} $\times$} &	{\color{green} \checkmark} &	{\color{blue} $\bullet$} &	{\color{green} \checkmark} \\ \midrule
\Large{Does the study address potential biases in data or analysis?} & {\color{green} \checkmark} &	{\color{blue} $\bullet$} &	{\color{green} \checkmark} &	{\color{green} \checkmark} &	{\color{green} \checkmark} &	{\color{blue} $\bullet$} &	{\color{green} \checkmark} &	{\color{green} \checkmark} &	{\color{blue} $\bullet$} &	{\color{green} \checkmark} &	{\color{green} \checkmark} &	{\color{green} \checkmark} &	{\color{green} \checkmark} &	{\color{green} \checkmark} &	{\color{green} \checkmark} &	{\color{green} \checkmark} &	{\color{green} \checkmark} &	{\color{green} \checkmark} &	{\color{green} \checkmark} &	{\color{green} \checkmark} &	{\color{green} \checkmark} &	{\color{green} \checkmark} &	{\color{green} \checkmark} &	{\color{green} \checkmark} &	{\color{blue} $\bullet$} &	{\color{green} \checkmark} \\ \midrule
\Large{Are the results clearly presented and supported by the data?} & {\color{green} \checkmark} &	{\color{blue} $\bullet$} &	{\color{green} \checkmark} &	{\color{green} \checkmark} &	{\color{green} \checkmark} &	{\color{green} \checkmark} &	{\color{green} \checkmark} &	{\color{green} \checkmark} &	{\color{green} \checkmark} &	{\color{green} \checkmark} &	{\color{green} \checkmark} &	{\color{green} \checkmark} &	{\color{green} \checkmark} &	{\color{green} \checkmark} &	{\color{green} \checkmark} &	{\color{green} \checkmark} &	{\color{green} \checkmark} &	{\color{green} \checkmark} &	{\color{green} \checkmark} &	{\color{green} \checkmark} &	{\color{green} \checkmark} &	{\color{green} \checkmark} &	{\color{green} \checkmark} &	{\color{green} \checkmark} &	{\color{green} \checkmark} &	{\color{green} \checkmark} \\ \midrule
\Large{Are limitations of the study acknowledged?} & {\color{green} \checkmark} &	{\color{green} \checkmark} &	{\color{green} \checkmark} &	{\color{green} \checkmark} &	{\color{green} \checkmark} &	{\color{green} \checkmark} &	{\color{green} \checkmark} &	{\color{green} \checkmark} &	{\color{green} \checkmark} &	{\color{green} \checkmark} &	{\color{green} \checkmark} &	{\color{green} \checkmark} &	{\color{green} \checkmark} &	{\color{green} \checkmark} &	{\color{green} \checkmark} &	{\color{green} \checkmark} &	{\color{green} \checkmark} &	{\color{green} \checkmark} &	{\color{green} \checkmark} &	{\color{green} \checkmark} &	{\color{green} \checkmark} &	{\color{green} \checkmark} &	{\color{green} \checkmark} &	{\color{green} \checkmark} &	{\color{green} \checkmark} &	{\color{green} \checkmark} \\ \midrule
\Large{Are the conclusions consistent with the results?} & {\color{green} \checkmark} &	{\color{green} \checkmark} &	{\color{green} \checkmark} &	{\color{green} \checkmark} &	{\color{green} \checkmark} &	{\color{green} \checkmark} &	{\color{green} \checkmark} &	{\color{green} \checkmark} &	{\color{green} \checkmark} &	{\color{green} \checkmark} &	{\color{green} \checkmark} &	{\color{green} \checkmark} &	{\color{green} \checkmark} &	{\color{green} \checkmark} &	{\color{green} \checkmark} &	{\color{green} \checkmark} &	{\color{green} \checkmark} &	{\color{green} \checkmark} &	{\color{green} \checkmark} &	{\color{green} \checkmark} &	{\color{green} \checkmark} &	{\color{green} \checkmark} &	{\color{green} \checkmark} &	{\color{green} \checkmark} &	{\color{green} \checkmark} &	{\color{green} \checkmark} \\ \midrule
\midrule
\textbf{\Large{Total `Y'}} & \textbf{10} & \textbf{6} & \textbf{10} & \textbf{9} & \textbf{9} & \textbf{7} & \textbf{9} & \textbf{8} & \textbf{8} & \textbf{9} & \textbf{10} & \textbf{8} & \textbf{9} & \textbf{9} & \textbf{9} & \textbf{10} & \textbf{9} & \textbf{10} & \textbf{9} & \textbf{10} & \textbf{10} & \textbf{7} & \textbf{9} & \textbf{10} & \textbf{8} & \textbf{10} \\
\midrule
\textbf{\Large{Risk Category}} & \Large{\textbf{$\downarrow$}} & \Large{\textbf{$\approx$}} & \Large{\textbf{$\downarrow$}} & \Large{\textbf{$\downarrow$}} & \Large{\textbf{$\downarrow$}} & \Large{\textbf{$\approx$}} & \Large{\textbf{$\downarrow$}} & \Large{\textbf{$\downarrow$}} & \Large{\textbf{$\downarrow$}} & \Large{\textbf{$\downarrow$}} & \Large{\textbf{$\downarrow$}} & \Large{\textbf{$\downarrow$}} & \Large{\textbf{$\downarrow$}} & \Large{\textbf{$\downarrow$}} & \Large{\textbf{$\downarrow$}} & \Large{\textbf{$\downarrow$}} & \Large{\textbf{$\downarrow$}} & \Large{\textbf{$\downarrow$}} & \Large{\textbf{$\downarrow$}} & \Large{\textbf{$\downarrow$}} & \Large{\textbf{$\downarrow$}} & \Large{\textbf{$\approx$}} & \Large{\textbf{$\downarrow$}} & \Large{\textbf{$\downarrow$}} & \Large{\textbf{$\downarrow$}} & \Large{\textbf{$\downarrow$}} \\
\bottomrule
\end{tabular}
}
\end{table}
The risk of bias was assessed for all 26 included studies using a simplified checklist as discussed in section \ref{study_bias_method}. The results of the risk of bias assessment are summarized in Table \ref{risk_bias_table}. Of all studies, 23 studies fall under low, 3 under moderate, and 0 under high risk of bias.

\subsection{Results of individual studies}
\label{results_of_individual_studies}
Tables \ref{summary_table} and \ref{summary_table_2} show data extraction results from individual studies. The studies reviewed span a range of topics and applications focusing on advancing automated methods in healthcare and analyzing treatment efficacy.

\subsection{Reporting biases}
\label{reprot_bias_reults}
Table \ref{risk_bias_table_2} provides the risk of bias assessment specific to missing results and reporting bias. A total of 9 studies were classified as low risk, 17 as moderate risk, and 0 as high risk. Several plausible biases were found for the studies included in this literature review, such as the lack of pre-registration for the study or the reference to a pre-registered protocol. This shows a gap in ensuring transparency in research design and execution. Additionally, many studies failed to provide access to supplementary materials, which limited the replicability and verification of reported findings. These findings emphasize the importance of adhering to open science practices to mitigate risks of reporting bias. These findings highlight variability in reporting practices, underscoring the need for consistent adherence to robust reporting standards in chronic pain research.
\begin{table}[h]
\centering
\caption{Risk of Reporting Bias and Missing Results.\newline{\color{green} \checkmark} = `Yes', {\color{red} $\times$} = `No', {\color{blue} $\bullet$} = `Unclear', {\textbf{$\downarrow$}} = `Low risk of bias', {\textbf{$\approx$}} = `Moderate risk of bias'.}
\label{risk_bias_table_2}
\renewcommand{\arraystretch}{1.8} 
\resizebox{\textwidth}{!}{%
\fontsize{25}{25}\selectfont 
\begin{tabular}{p{7cm}llllllllllllllllllllllllll}
\toprule
 & \Large{\cite{necaise_peer_2024}} & \Large{\cite{lotz_exploration_2023}} & \Large{\cite{nunes_modeling_2023}} & \Large{\cite{chaturvedi_distributions_2024}} & \Large{\cite{venerito_large_2024}} & \Large{\cite{j_m_reinen_remotely-captured_2024}} & \Large{\cite{sarker_chronicpain_2023}} & \Large{\cite{himmelstein_examination_2022}} & \Large{\cite{c_agurto_exploring_2024}} & \Large{\cite{nunes_chronic_2023}} & \Large{\cite{aggarwal_cross-modal_2023}} & \Large{\cite{gordon_relationship_2023}} & \Large{\cite{ashar_yk_reattribution_2023}} & \Large{\cite{dobscha_mental_2023}} & \Large{\cite{goldstein_characterizing_2023}} & \Large{\cite{schirle_two_2021}} & \Large{\cite{chaturvedi_development_2023}} & \Large{\cite{chen_understanding_2019}} & \Large{\cite{hylan_automated_2015}} & \Large{\cite{berger_quantitative_2021}} & \Large{\cite{taylor_use_2019}} & \Large{\cite{chaturvedi_development_2021}} & \Large{\cite{davidson_job-related_2021}} & \Large{\cite{branco_predicting_2022}} & \Large{\cite{goudman_social_2022}} & \Large{\cite{yang_combining_2020}} \\ 
\midrule
\Large{Are all datasets described in the methods section also reported in the results section?} & {\color{green} \checkmark} &	{\color{green} \checkmark} &	{\color{green} \checkmark} &	{\color{red} $\times$} &	{\color{green} \checkmark} &	{\color{green} \checkmark} &	{\color{green} \checkmark} &	{\color{green} \checkmark} &	{\color{green} \checkmark} &	{\color{green} \checkmark} &	{\color{green} \checkmark} &	{\color{green} \checkmark} &	{\color{green} \checkmark} &	{\color{green} \checkmark} &	{\color{green} \checkmark} &	{\color{green} \checkmark} &	{\color{green} \checkmark} &	{\color{green} \checkmark} &	{\color{green} \checkmark} &	{\color{green} \checkmark} &	{\color{green} \checkmark} &	{\color{green} \checkmark} &	{\color{green} \checkmark} &	{\color{green} \checkmark} &	{\color{green} \checkmark} &	{\color{green} \checkmark} \\
\midrule
\Large{Are all evaluation metrics listed in the methods section reported in the results?} &	{\color{green} \checkmark} &	{\color{red} $\times$} &	{\color{red} $\times$} &	{\color{red} $\times$} &	{\color{green} \checkmark} &	{\color{green} \checkmark} &	{\color{red} $\times$} &	{\color{blue} $\bullet$} &	{\color{green} \checkmark} &	{\color{green} \checkmark} &	{\color{green} \checkmark} &	{\color{green} \checkmark} &	{\color{green} \checkmark} &	{\color{red} $\times$} &	{\color{green} \checkmark} &	{\color{green} \checkmark} &	{\color{red} $\times$} &	{\color{green} \checkmark} &	{\color{green} \checkmark} &	{\color{red} $\times$} &	{\color{red} $\times$} &	{\color{red} $\times$} &	{\color{red} $\times$} &	{\color{green} \checkmark} &	{\color{green} \checkmark} &	{\color{green} \checkmark}\\ \midrule
\Large{Does the study report results for all intended tasks or objectives?} &	{\color{green} \checkmark} &	{\color{blue} $\bullet$} &	{\color{green} \checkmark} &	{\color{green} \checkmark} &	{\color{green} \checkmark} &	{\color{green} \checkmark} &	{\color{red} $\times$} &	{\color{green} \checkmark} &	{\color{green} \checkmark} &	{\color{green} \checkmark} &	{\color{green} \checkmark} &	{\color{green} \checkmark} &	{\color{green} \checkmark} &	{\color{green} \checkmark} &	{\color{green} \checkmark} &	{\color{green} \checkmark} &	{\color{red} $\times$} &	{\color{green} \checkmark} &	{\color{green} \checkmark} &	{\color{green} \checkmark} &	{\color{green} \checkmark} &	{\color{green} \checkmark} &	{\color{green} \checkmark} &	{\color{red} $\times$} &	{\color{green} \checkmark} &	{\color{green} \checkmark}\\
\midrule
\Large{Are the results for negative or null findings explicitly reported?} & {\color{green} \checkmark} &	{\color{blue} $\bullet$} &	{\color{green} \checkmark} &	{\color{green} \checkmark} &	{\color{green} \checkmark} &	{\color{green} \checkmark} &	{\color{green} \checkmark} &	{\color{green} \checkmark} &	{\color{green} \checkmark} &	{\color{green} \checkmark} &	{\color{green} \checkmark} &	{\color{green} \checkmark} &	{\color{green} \checkmark} &	{\color{green} \checkmark} &	{\color{green} \checkmark} &	{\color{green} \checkmark} &	{\color{green} \checkmark} &	{\color{green} \checkmark} &	{\color{green} \checkmark} &	{\color{green} \checkmark} &	{\color{blue} $\bullet$} &	{\color{green} \checkmark} &	{\color{green} \checkmark} &	{\color{green} \checkmark} &	{\color{green} \checkmark} &	{\color{red} $\times$} \\ 
\midrule
\Large{Is there consistency between the study’s objectives and its results?} & {\color{green} \checkmark} &	{\color{green} \checkmark} &	{\color{green} \checkmark} &	{\color{green} \checkmark} &	{\color{green} \checkmark} &	{\color{green} \checkmark} &	{\color{green} \checkmark} &	{\color{green} \checkmark} &	{\color{green} \checkmark} &	{\color{green} \checkmark} &	{\color{green} \checkmark} &	{\color{green} \checkmark} &	{\color{green} \checkmark} &	{\color{green} \checkmark} &	{\color{green} \checkmark} &	{\color{green} \checkmark} &	{\color{green} \checkmark} &	{\color{green} \checkmark} &	{\color{green} \checkmark} &	{\color{green} \checkmark} &	{\color{green} \checkmark} &	{\color{green} \checkmark} &	{\color{green} \checkmark} &	{\color{green} \checkmark} &	{\color{green} \checkmark} &	{\color{green} \checkmark}
 \\ \midrule
\Large{Does the study acknowledge any missing results or limitations in reporting?} & {\color{blue} $\bullet$} &	{\color{green} \checkmark} &	{\color{blue} $\bullet$} &	{\color{green} \checkmark} &	{\color{green} \checkmark} &	{\color{green} \checkmark} &	{\color{green} \checkmark} &	{\color{green} \checkmark} &	{\color{green} \checkmark} &	{\color{green} \checkmark} &	{\color{green} \checkmark} &	{\color{green} \checkmark} &	{\color{green} \checkmark} &	{\color{green} \checkmark} &	{\color{green} \checkmark} &	{\color{green} \checkmark} &	{\color{green} \checkmark} &	{\color{green} \checkmark} &	{\color{green} \checkmark} &	{\color{green} \checkmark} &	{\color{green} \checkmark} &	{\color{green} \checkmark} &	{\color{green} \checkmark} &	{\color{green} \checkmark} &	{\color{green} \checkmark} &	{\color{green} \checkmark}
 \\ \midrule
\Large{Was the study pre-registered, or does it reference a pre-registered protocol?} &	{\color{red} $\times$} &	{\color{red} $\times$} &	{\color{red} $\times$} &	{\color{red} $\times$} &	{\color{red} $\times$} &	{\color{green} \checkmark} &	{\color{red} $\times$} &	{\color{red} $\times$} &	{\color{green} \checkmark} &	{\color{red} $\times$} &	{\color{red} $\times$} &	{\color{red} $\times$} &	{\color{green} \checkmark} &	{\color{red} $\times$} &	{\color{red} $\times$} &	{\color{red} $\times$} &	{\color{red} $\times$} &	{\color{blue} $\bullet$} &	{\color{red} $\times$} &	{\color{green} \checkmark} &	{\color{red} $\times$} &	{\color{red} $\times$} &	{\color{red} $\times$} &	{\color{green} \checkmark} &	{\color{red} $\times$} &	{\color{red} $\times$}
 \\ \midrule

\Large{Does the study provide access to supplementary material or raw data?} &	{\color{red} $\times$} &	{\color{green} \checkmark} &	{\color{blue} $\bullet$} &	{\color{blue} $\bullet$} &	{\color{green} \checkmark} &	{\color{red} $\times$} &	{\color{green} \checkmark} &	{\color{green} \checkmark} &	{\color{red} $\times$} &	{\color{red} $\times$} &	{\color{green} \checkmark} &	{\color{red} $\times$} &	{\color{blue} $\bullet$} &	{\color{green} \checkmark} &	{\color{green} \checkmark} &	{\color{green} \checkmark} &	{\color{green} \checkmark} &	{\color{red} $\times$} &	{\color{red} $\times$} &	{\color{green} \checkmark} &	{\color{red} $\times$} &	{\color{green} \checkmark} &	{\color{red} $\times$} &	{\color{green} \checkmark} &	{\color{blue} $\bullet$} &	{\color{red} $\times$}
\\ \midrule
\midrule
\textbf{\Large{Total `Y'}} & \textbf{5} &	\textbf{4} &	\textbf{4} &	\textbf{4} &	\textbf{7} &	\textbf{7} &	\textbf{5} &	\textbf{6} &	\textbf{7} &	\textbf{6} &	\textbf{7} &	\textbf{6} &	\textbf{7} &	\textbf{6} &	\textbf{7} &	\textbf{7} &	\textbf{5} &	\textbf{6} &	\textbf{6} &	\textbf{7} &	\textbf{4} &	\textbf{6} &	\textbf{5} &	\textbf{7} &	\textbf{6} &	\textbf{5}
 \\ 
 \midrule
\textbf{\Large{Risk Category}} & \Large{\textbf{$\approx$}} & \Large{\textbf{$\approx$}} & \Large{\textbf{$\approx$}} & \Large{\textbf{$\approx$}} & \Large{\textbf{$\downarrow$}} & \Large{\textbf{$\downarrow$}} & \Large{\textbf{$\approx$}} & \Large{\textbf{$\approx$}} & \Large{\textbf{$\downarrow$}} & \Large{\textbf{$\approx$}} & \Large{\textbf{$\downarrow$}} & \Large{\textbf{$\approx$}} & \Large{\textbf{$\downarrow$}} & \Large{\textbf{$\approx$}} & \Large{\textbf{$\downarrow$}} & \Large{\textbf{$\downarrow$}} & \Large{\textbf{$\approx$}} & \Large{\textbf{$\approx$}} & \Large{\textbf{$\approx$}} & \Large{\textbf{$\downarrow$}} & \Large{\textbf{$\approx$}}& \Large{\textbf{$\approx$}} & \Large{\textbf{$\approx$}} & \Large{\textbf{$\downarrow$}} & \Large{\textbf{$\approx$}} & \Large{\textbf{$\approx$}}
 \\ 
\bottomrule
\end{tabular}
}
\end{table}

\begin{sidewaystable}
\vspace{15cm}
\centering
\caption{Summary of the 26 included Studies.}
\label{summary_table}
\renewcommand{\arraystretch}{1.7}
\resizebox{\textwidth}{!}{
\begin{tabular}{p{1cm} p{2.5cm} p{6cm} p{8cm} p{6cm} p{3cm} p{4cm} p{7cm}}
\hline
\textbf{Year\newline Ref.} & \textbf{Study Design} & \textbf{Research Question} & \textbf{Dataset} & \textbf{NLP Tech.} & \textbf{Participant Count} & \textbf{Age (Years)} & \textbf{Results} \\ \hline
2015 \cite{hylan_automated_2015} & 	Observational, longitudinal & 	Can EHR data predict problem opioid use in chronic opioid therapy? & 	2,752 noncancer pain patients from Group Health Cooperative (2008–2010) & 	Regular Expression & 	2,752 \newline(train: 1,449 \& Val: 1,303) & 	18+ & 	C-statistic: 0.739 (95\% CI = .688, .790); Sensitivity: 60.1\%; Specificity: 71.6\%; PPV: 11.4\%; NPV: 96.7\% \\ \hline
2019 \cite{chen_understanding_2019} & 	Observational, Secondary data analysis & 	How text mining \& visual analytics reveal engagement in adolescent pain interventions? & 	123 adolescents with chronic idiopathic pain \& their parents in an internet-based CBT intervention & 	LDA\newline k-means clustering & 	273 & 	11-17 & 	Participants' messages acc: 76.5\%, Coaches' messages acc: 99.3\% \\ \hline
2019 \cite{taylor_use_2019} & 	retrospective cohort, observational & 	What are the frequency and predictors of CIH therapy use in veterans with chronic pain? & 	Retrospective cohort of 530,216 younger veterans with chronic MSK pain (2010–2013, VHA) & 	SVM & 	530,216 & 	18-40 & 	AUC ranged from 82.8\% to 91.8\%, Accuracy ranged from 79.8\% to 85.8\% \\ \hline
2020 \cite{yang_combining_2020} & 	Observational, retrospective & 	How can DL \& NLP improve disorder classification from discharge summaries? & 	MIMIC-III & 	Word and sentence-level CNN, SVM with id-itf \& word embedding & 	1610 & 	N/A & 	The averaged metrics are Precision: 82.30\%, Recall: 20.72\%, ROCAUC: 60.00\%, and F1: 31.89\% \\ \hline
2021 \cite{berger_quantitative_2021} & 	Randomized, double-blind, placebo-controlled & Can patient discourse features identify placebo responders in back pain? & 	Exit interviews from 66 chronic back pain patients (NCT02013427) & 	Logistic Regression, Linear SVM, L1 (lasso) regularization, L2 (ridge) regularization & 	125 & 18+\newline Men Avg.: 48.8±1.9 \& Women Avg. 41.84±2.7 & 	Accuracy: 79\%, Sensitivity: 0.82, Specificity: 0.75 (SVM with L1 regularization) \\ \hline
2021 \cite{davidson_job-related_2021} & Retrospective & 	What job issues do nurses face before suicide? & 	NVDRS (2003–2017, CDC) data on 203 nurse suicides & 	LDA, Treebank Word Tokenizer, WordNet Lemmatizer & 	203 nurses & 	21+ & 	N/A \\ \hline
2021 \cite{chaturvedi_development_2021} & 	Observational, exploratory analysis & 	How is pain described, \& which texts aid in creating a pain lexicon? & CRIS, MIMIC-III, Twitter (`chronic pain' tweets), Reddit (chronic pain subreddit) & 	word2vec, FastText & 	200 documents\newline (50 per dataset) & 	N/A & 	N/A \\ \hline
2021 \cite{schirle_two_2021} & 	Retrospective observational pilot & 	How can data-driven methods in EHRs identify problematic opioid use? & 	Vanderbilt's BioVU biorepository & 	Text-based Score using ScispaCy & 	29868 & 20+ & 	AUC for Text-based score: 0.79, AUC for Comorbidity score: 0.76 \\ \hline
2022 \cite{goudman_social_2022} & 	Observational & 	What do chronic pain patients discuss on Reddit? & 	937 text posts from `r/ChronicPain' subreddit (collected via PRAW on 6 Sept 2021) &	LDA\newline Bag-of-words model & 709 unique authors & 	N/A & 	N/A \\ \hline
2022 \cite{himmelstein_examination_2022} & 	cross-sectional observational & How often does stigmatizing language in admission notes vary by condition and race? & 48651 admission notes (2018) from a large urban medical center & Tokenization into unigrams \& bigrams, Porter2 stemming algorithm & 	29,783 patients\newline 1,932 clinicians & 	Mean: 46.9\newline SD: 27.6 & 	N/A \\ \hline
2022 \cite{branco_predicting_2022} & Double-blind, placebo-controlled & 	Can language predict placebo response and distinguish drug from placebo responders? & Two RCTs (NCT02013427) on chronic back pain w/ language interviews & Linguistic Inquiry Word Count\newline Semantic proximity & Study 1: 66\newline Study 2: 50 & Mean: 45.3\newline SD: 2.3 &	Placebo arm: AUC 0.708, F1 0.65, Precision 0.68, Recall 0.65, Balanced Acc. 67\% \\ \hline
2023 \cite{ashar_yk_reattribution_2023} & randomized, placebo-controlled & 	Does reattributing back pain to mind/brain processes relieve pain in PRT? & Chronic back pain adults responses on pain causes (NCT03294148, 2017–2019) & 	Text scaling (LSA/PCA)\newline Rule-based (keywords matching) & 151 & 21-70 & Cohen $\kappa$ at:\newline pretreatment: 0.42\newline posttreatment: 0.68 \\ \hline
2023 \cite{chaturvedi_development_2023} & Observational, retrospective & How can we create a labeled pain corpus from mental health EHRs for NLP? & 	CRIS\newline A manually annotated corpus of pain mentions in mental health EHRs & KNN\newline SVM\newline BERT & 723 patients & $\leq$20: 10\%, 21-40: 21\%\newline 41-60: 25\%, 61-80: 33\%, $>$80: 10\% & 	F$_1$: 0.98\newline (95\% CI 0.98-0.99) \\ \hline
2023 \cite{nunes_modeling_2023} & 	Observational & How can the RRCP dataset model chronic pain experiences on Reddit? & Reddit Reports of Chronic Pain (RRCP)\newline 86537 submissions (2013–2020) from 12 chronic pain-related subreddits & K-Means\newline LDA\newline VADER sentiment analysis & 	44,815 unique authors & 13+ & 	N/A \\ \hline

\end{tabular}
}
\end{sidewaystable}

\begin{sidewaystable}
\vspace{15cm}
\centering
\caption{Summary of the 26 included studies, cont'd.}
\label{summary_table_2}
\renewcommand{\arraystretch}{1.7}
\resizebox{\textwidth}{!}{
\begin{tabular}{p{1cm} p{2.5cm} p{6cm} p{8cm} p{6cm} p{3cm} p{4cm} p{7cm}}
\hline
\textbf{Year\newline Ref.} & \textbf{Study Design} & \textbf{Research Question} & \textbf{Dataset} & \textbf{NLP Tech.} & \textbf{Participant Count} & \textbf{Age (Years)} & \textbf{Results} \\ \hline
2023 \cite{dobscha_mental_2023} & 	retrospective cohort, observational & 	Are mental health conditions linked to pain care quality in VHA primary care? & VHA CDW linked to MSD cohort w/ EHR data from 134,508 visits with moderate/severe MSK pain (2017) & 	Rule-based NLP & 	62,721 Veterans & 	Avg.: 50.5 yrs\newline SD: 16.6 & 	F-measure: 0.92 \\ \hline
2023 \cite{goldstein_characterizing_2023} & 	Observational, retrospective & 	What circumstances precede female firearm suicide in coroner \& police reports? & 	NVDRS (2014–2018) data on female firearm suicides (40 states, Puerto Rico) &	Naive Bayes, Random Forest\newline SVM, Gradient Boosting\newline Lemmatization, TF-IDF & 	1462 & 	Mean: 47\newline SD: 6.9 & 	SVM: Specificity 98.1\%, Sensitivity 70.5\%, F\textsubscript{1} 79.1, PPV 90.5\%, AUC-ROC 84.3\%\newline GB: Specificity 98.7\%, PPV 93.2\%. \\ \hline
2023 \cite{gordon_relationship_2023} & 	Retrospective cross-sectional & 	Do pain screening and reports differ between LGBT and non-LGBT Veterans? & 	VHA CDW (FY 2010–2019), covering over 9 million Veterans & 	Keywords based matching (Regex) & 	1,149,486 (LGBT: 218,154) & 10-year age groups: $<$25 to $\geq$75 & 	Sensitivity: 88.2\%\newline Specificity: 91.5\% \\ \hline
2023 \cite{nunes_chronic_2023} & 	Observational, single-site, proof-of-concept & 	How can NLP estimate pain intensity from patient narratives? & 	Interviews with chronic pain patients from Centro Hospitalar Universitário de São João (Oct 2019–Oct 2020) & 	TF-IDF, POS TF-IDF, SVM, DT & 	94 & Avg: 56.4 ± 12.7 & 	SVM F\textsubscript{1}: 0.60\newline Classification acc.:\newline 92\% mild, 42\% moderate, 13\% severe \\ \hline
2023 \cite{aggarwal_cross-modal_2023} & 	Observational, retrospective, stratified analysis & 	Does Twitter language predict community pain and its variations beyond demographics? & Gallup-Sharecare Well-Being Index\newline CTLB ($\sim$1.5B county-level tweets)\newline $\sim$1.8M pain-related tweets matched to Gallup counties & 	Extra Tree Classifier\newline RF\newline Content Specific LDA & 	2,541,688 responses & 	N/A & 	RF: AUC 0.86, F\textsubscript{1} 0.64, Precision 0.85, Recall 0.61\newline ETC: AUC 0.86, F\textsubscript{1} 0.66, Precision 0.84, Recall 0.62 \\ \hline
2023 \cite{lotz_exploration_2023} & 	exploratory methodological & How can LLMs, SGs, \& KGs synthesize literature to aid chronic back pain research? & 83 LBP literature & 	GPT3 \& GPT4\newline SGs, KGs & 	N/A & 	N/A & 	Acc: 60\%; F\textsubscript{1}: 0.70 (biomechanics), 0.70 (psychology), 0.36 (both), 0.46 (neither) \\ \hline
2023 \cite{sarker_chronicpain_2023} & 	Observational study & 	How can Twitter reveal chronic pain experiences \& alternative therapies? & 	4,998 annotated tweets with chronic pain hashtags/keywords, labeled for self-reported chronic pain experiences & 	RoBERTa, SciBERT\newline BioClinicalBERT, BERTweet\newline BioBERT, Clinical\_KB\_BERT & 	22,795 Twitter users & 	N/A & 	RoBERTa F\textsubscript{1}: 0.84\newline (95\% CI: 0.80–0.89) \\ \hline

2024 \cite{venerito_large_2024} & 	retrospective observational & Can LLM sentiment analysis aid fibromyalgia (FM) diagnosis via pain patterns? & 	Patients with and without FM (per 2016 ACR criteria) chronic pain from a rheumatology clinic (Jan 8–20, 2024). & Mistral-7B-Instruct-v0.2\newline HuggingFace Transformers API for attention weight analysis & 80 (FM: 40, non-FM pain controls: 40) & 	$\sim$38.76 to 57.68 & 	Prompt-engineered:\newline Acc. 0.87, Precision 0.92, Recall 0.84, Specificity 0.82, AUROC 0.86 \\ \hline
2024 \cite{necaise_peer_2024} & 	Observational, retrospective & 	Do online peer support users align their pain language with others? & 	Reddit users' comment histories in a chronic pain support community (10 years, 68,206 interactions) with $\geq$100 posts per user & Linguistic Inquiry and Word Count (LIWC) dictionary\newline VADER & 199 users & 	N/A & 	t(198)=4.02, $p<.001$, d=0.40 (common power)\\ \hline
2024 \cite{chaturvedi_distributions_2024} & 	observational, cross-sectional, retrospective & 	What are the demographics and diagnoses of pain in mental health records, and how do they overlap in care? & CRIS\newline Lambeth DataNet (primary care records), linked for patient-level analysis across care settings & 	SapBERT\newline MedCATTrainer for NER & 	27211 & 	18+\newline Mean: 44\newline IQR: 29-55 & 	F\textsubscript{1} for:\newline Pain classification: 0.98\newline Anatomy identification: 0.94 \\ \hline
2024 \cite{j_m_reinen_remotely-captured_2024} & longitudinal, observational & How can sentiment analysis track health in chronic pain patients? & 	ENVISION Study (NCT03240588) on chronic back/leg pain: 34,447 free-text samples (Nov 2019–2023) & 	Sentence Transformers, Watson NLU, k-means clustering\newline Hugging Face hub Transformers & 	242 & 	N/A & 	N/A \\ \hline
2024 \cite{c_agurto_exploring_2024} & 	Prospective, observational & How can text and audio from chronic pain patients infer well-being? & 	Text \& voice responses from chronic pain patients in NAVITAS \& ENVISION studies (Dec 2019–July 2023, 30 U.S. sites) & 	RoBERTa, GeMAPS v2.0, Whisper Open AI, Lasso regression, Ridge regression, SVM & 	166 & 	Text responses: 49.6-73.2; Voice responses: 52-74 & 	$\rho$: Mood (text) r=0.46, Pain (text) r=0.44, Sleep (text) r=0.40, Pain (voice) r=0.45, Alertness (voice) r=0.42 \\ \hline
\end{tabular}
}
\end{sidewaystable}

\section{Discussion}
\subsection{Central Findings}
\label{central_findings}
\subsubsection{Methodological Approach}: Based on the results tables \ref{summary_table} \& \ref{summary_table_2} provided, it is clear that the integration of NLP techniques in healthcare, particularly in chronic pain research, has evolved significantly over time. Early studies (e.g., 2015, 2019) often relied on rule-based methods like regular expressions and simple machine learning models such as SVMs, while more recent research has shifted toward advanced deep learning techniques, including BERT \cite{devlin_bert_2019} and RoBERTa \cite{liu_roberta_2019}. This evolution reflects a growing sophistication in the field that is leveraging larger datasets, such as MIMIC-III \cite{noauthor_mimic-iii_nodate} and VHA, and unstructured sources like Reddit and Twitter to gain deeper insights into patient experiences.

The findings across studies reveal promising results in using automated and computational methods for health analysis. For example, the use of NLP in \citet{hylan_automated_2015} allowed for improved surveillance of opioid use which showcases the potential of data-driven solutions in addressing specific health crises. Another key insight is the effectiveness of machine learning models, such as the ws-CNN used in \citet{yang_combining_2020}, which enhances classification accuracy in patient phenotyping. Despite different intervention types, a recurring outcome is the growing evidence supporting automated methods for healthcare, highlighting increased accuracy and predictive capacity as consistent themes across the papers.

\subsubsection{Types of Research Problems}: Many studies aim to predict health outcomes, such as opioid misuse \cite{hylan_automated_2015}, placebo responders \cite{berger_quantitative_2021} or fibromyalgia diagnosis. These studies emphasize classification metrics (AUC, F\textsubscript{1}-score) to validate their models' effectiveness. However, a significant subset of studies \cite{goudman_social_2022, davidson_job-related_2021} explores patient narratives to understand broader themes, such as the language of chronic pain on Reddit or the factors contributing to nurse suicides. These studies are less focused on prediction and more on identifying patterns or building resources, such as pain lexicons. Finally, studies targeting specific populations, such as \citet{chen_understanding_2019} for adolescents or \citet{gordon_relationship_2023} for LGBT veterans, tailor their research to demographic or population-specific datasets, leading to unique methodological adaptations.

\subsubsection{Data Sources}: The studies highlight a diverse range of data sources, emphasizing the distinction between structured and unstructured datasets. Structured clinical datasets, such as MIMIC-III \cite{yang_combining_2020} and Vanderbilt’s BioVU \cite{schirle_two_2021}, are often used for predictive analytics due to their predefined fields and compatibility with rule-based or feature-based methods. Conversely, unstructured data sources like Reddit \cite{nunes_modeling_2023, goudman_social_2022} and Twitter \cite{sarker_chronicpain_2023} require advanced NLP techniques to process free-text narratives effectively, enabling insights into public discourse and patient experiences. The scale of datasets also varies widely; large-scale clinical datasets, such as those used in \cite{taylor_use_2019} with over 530,000 patients, support robust model training, while smaller datasets, such as the 66 participants in \cite{branco_predicting_2022}, offer qualitative depth but face limitations in generalizability.

\subsubsection{Large Language Model (LLM)}:  \citet{lotz_exploration_2023} explored GPT-3 and GPT-4 for synthesizing chronic back pain literature, categorizing findings across biomechanics and psychology, though modest F\textsubscript{1} (0.46 for combined domains) revealed challenges in interdisciplinary understanding. \citet{venerito_large_2024} used prompt-engineered sentiment analysis with Mistral-7B for fibromyalgia diagnosis, achieving high accuracy (0.87) but limited to a single application. Similarly, \citet{c_agurto_exploring_2024} combined textual and audio data using RoBERTa and Whisper models, identifying correlations between mood, pain, and alertness but struggling to integrate these insights into clinical workflows. The number of studies utilizing LLMs in this review is limited. However, notably, all three studies involving LLMs were published in 2024, which indicates that their adoption in the field is gaining traction.
\subsection{Research Gap}
\label{limitation}
\subsubsection{Dataset size \& diversity}: One pervasive issue is the lack of validation and generalization of findings across diverse populations. \citet{venerito_large_2024} highlighted the need for extensive validation in large-scale studies to confirm the reliability of observed outcomes. This aligns with earlier observations by \citet{yang_combining_2020}, who pointed to inadequate sample sizes that undermine the statistical power and generalizability of many studies. 
The representation of diverse populations in studies remains insufficient, posing a challenge to the equity and applicability of findings. Studies such as those by \citet{taylor_use_2019} and \citet{chaturvedi_distributions_2024} emphasized the need to include diverse demographic, cultural, and socioeconomic groups. Without this inclusivity, research risks producing interventions that fail to address the needs of underserved populations, perpetuating health disparities.

\subsubsection{Context-specific insights}: Mechanistic understanding and context-specific insights also remain underexplored. For instance, \citet{chen_understanding_2019} stressed the importance of identifying the mechanisms driving intervention efficacy, including the “how” and “why” of their success in specific contexts. \citet{necaise_peer_2024} further underscored the role of social dynamics in shaping intervention outcomes, which has been largely overlooked. Similarly, \citet{chaturvedi_distributions_2024} highlighted the challenge of differentiating interventions that actively influence outcomes from those merely correlated with behavioral changes. This lack of clarity in mechanisms calls for future studies that delve deeper into contextual and cultural nuances, ensuring that interventions are both targeted and effective for specific subgroups.

\subsubsection{Standard evaluation metrics}: Measurement challenges present another significant barrier. \citet{berger_quantitative_2021} pointed to difficulties in standardizing metrics, such as those for placebo responses in clinical trials, which complicate cross-study comparisons. \citet{j_m_reinen_remotely-captured_2024} emphasized the need to assess the balance of positive versus negative inputs in digital health interventions, a largely neglected area. Furthermore, \citet{c_agurto_exploring_2024} highlighted the potential of unstructured speech data in chronic pain research, yet its integration into clinical studies remains limited due to a lack of methodological tools. Overcoming these challenges requires the development of validated, standardized measurement techniques that enhance consistency across diverse research contexts.

\subsubsection{Reproducability}:
It has been found that out of all the studies reviewed, only three open-sourced their analysis code-base for other researchers to utilize \cite{nunes_modeling_2023,chaturvedi_distributions_2024,chaturvedi_development_2021}. This highlights a significant gap in reproducibility and the broader accessibility of methodological advancements.

\subsection{Limitations}
\label{review_limitation}
While the systematic review employed a robust methodology to ensure comprehensive coverage, several limitations should be acknowledged. The reliance on specific databases (PubMed, Web of Science, IEEE Xplore, Scopus, and ACL Anthology) may have excluded relevant studies published in other niche or non-indexed sources, potentially introducing a selection bias. Additionally, the exclusion of certain publication types, such as preprints and gray literature, might have omitted emerging or unconventional perspectives that are not yet peer-reviewed \cite{adams_searching_2016}. Although backward and forward citation chasing using Citationchaser aimed to minimize gaps, this process depends on the completeness and accuracy of citation networks, which can vary across studies. Furthermore, restricting the search to articles published between January 2014 and September 2024 may have excluded foundational work or older studies that remain pertinent to the field. Finally, the manual screening of studies (despite following predefined inclusion criteria) is inherently subjective and could be influenced by the reviewers’ interpretation, which underscores the need for critical appraisal in systematic reviews. These limitations should be considered when interpreting the findings of this review.

\subsection{Future Research Directions}
\label{future_research}
\subsubsection{Cross-institute validation}: NLP has demonstrated its utility as an end-to-end process
for classifying and identifying class labels using chronic pain-related textual dataset(s). It will continue to be used for solving more chronic pain-related research questions and is likely to be used as a default approach
along with many other models (as discussed in Tables \ref{summary_table} \& \ref{summary_table_2})
giving the best reliability. Overall, the body of evidence underscores the need for methodological rigor, innovative measurement approaches, and interdisciplinary collaboration. Validation through large-scale inter-university/hospital studies and a stronger focus on contextual and cultural sensitivity are critical priorities for future research. Addressing these gaps will not only strengthen the theoretical foundations of intervention design but also ensure practical applications that are inclusive and impactful.

\subsubsection{Promoting Transparency and Accessibility}: Future researchers should make an attempt to publicize their code on popularly used version-controlling platforms such as Github, Zenodo and others. Only 11\% of the studies in our review have shared code base \cite{nunes_modeling_2023,chaturvedi_distributions_2024,chaturvedi_development_2021}. Sharing the codebase will not only allow fellow researchers to
reproduce the models and contribute in as many ways as possible but also ensure the reliability of the results presented in the study. Including details like training and test datasets, code to
generate the model, dependencies used, parameters used, and
the computational capacity used to train the model would help other researchers improve upon the existing models.

\subsubsection{Addressing Cross-Linguistic}: The results demonstrate that advanced methods generally achieve higher precision and recall, with some models, like RoBERTa and semantic proximity approaches, achieving F\textsubscript{1} scores above 0.8. However, certain gaps remain, such as limited attention to cross-linguistic and multimodal datasets and the underrepresentation of non-English narratives. Future research studies should also focus on validating existing (or developing new) models on non-English languages. One potential solution could be training LLMs on diverse cultural and demographic data to create domain-specific corpora that address gaps in socioeconomic, geographic, and linguistic diversity.

\subsubsection{LLM in future studies}: Future research could enhance LLMs by refining multimodal integration (e.g., text and audio), improving literature synthesis with domain-specific training, and ensuring generalizability across chronic pain conditions using larger, diverse datasets. Moreover, dynamic real-time personalization and ethical applications, including bias detection, remain critical areas for advancing LLM use in chronic pain research, bridging current gaps and expanding their clinical impact.

\section{Conclusions}
This systematic review highlights the evidence related to NLP systems for chronic pain. The review demonstrates NLP's utility across various applications such as chronic pain data classification, patient discourse analysis, and treatment outcome prediction. While significant progress has been achieved, several challenges remain, including the need for diverse datasets, standard evaluation metrics, and reproducible research practices. Future efforts should focus on addressing these gaps through interdisciplinary collaboration, methodological rigor, and inclusivity to ensure impactful, equitable, and replicable solutions. By leveraging advancements in NLP, the field can continue to drive innovation in chronic pain management and research.

\section*{Acknowledgments} 
Thanks to Prof. Clifford for sharing his insights on writing a decent scientific article that inspired this report. Many thanks to Prof. Reyna for the in-class discussions. Finally, thanks, Tim \& Masoud, for suggesting improvements on this paper.
\section*{Conflict of Interest}
\label{coi_statement}
The author (SR) has no conflicts of interest to disclose.

\section*{Funding}
\label{support}
None.
\def\newblock{\hskip .11em plus .33em minus .07em}


\clearpage
\bibliographystyle{unsrtnat}
\bibliography{references}

\newpage
\markboth{\small APPENDIX}{\small APPENDIX}  

\appendix
\section{List of Acronyms}
Table \ref{acronyms} lists the acronyms and abbreviations used in this
paper.
\begin{table}[h!]
\centering
\caption{List of Acronyms}
\label{acronyms}
\renewcommand{\arraystretch}{1.5}
\resizebox{\textwidth}{!}{%
\begin{tabular}{p{2.5cm}p{7cm}p{2.5cm}p{7cm}p{2.5cm}p{7cm}}
\toprule
\textbf{Acronym} & \textbf{Definition} & \textbf{Acronym} & \textbf{Definition} & \textbf{Acronym} & \textbf{Definition} \\
\midrule
CRIS & Clinical Record Interactive Search & EHR & Electronic Health Records & MIMIC & Medical Information Mart for Intensive Care \\
GeMAPS & Geneva Minimalistic Acoustic Parameter Set & LDA & Latent Dirichlet Allocation & CBT & Cognitive behavioral therapy \\
CIH & Complementary and Integrative Health & VHA & Veterans Health Administration & NVDRS & National Violent Death Reporting System\\
CNN & Convolutional Neural Network & VADER & Valence Aware Dictionary and Sentiment Reasoner & CDW & Corporate Data Warehouse \\
TF-IDF & Term frequency-inverse document frequency & LBP & Lower Back Pain & GPT & Generative Pre-trained Transformer \\

\bottomrule
\end{tabular}
}
\end{table}
\section{Search Query Results}
\label{search_results}
\begin{table}[htbp]

    \centering
    \caption{Search Strategy Results}
    \renewcommand{\arraystretch}{1.5} 
    \setlength{\tabcolsep}{7pt} 

    \begin{tabular}{|c|l|c|c|}
        \hline
        \textbf{S.No.} & \textbf{Database Name} & \textbf{Search Date} & \textbf{Number of Records} \\ \hline
        1 & PubMed & 09/15/2024 & 08 \\ \hline
        2 & Web of Science & 09/15/2024 & 36\\ \hline
        3 & IEEE Xplore & 09/15/2024 & 06\\ \hline
        4 & SCOPUS & 09/15/2024 & 45\\ \hline
        5 & ACL Anthology & 09/15/2024 & 32\\ \hline
        \multicolumn{3}{|r|}{\textbf{Total}} & \textbf{127} \\ \hline
    \end{tabular}
    \label{search_results_tab}
\end{table}

\section{Search Strategy}
\label{search_strategy_app}
\subsection{PubMed}
(chronic pain) AND ((natural language processing) OR (large language model)) Filters: Adaptive Clinical Trial, Books and Documents, Case Reports, Classical Article, Clinical Conference, Clinical Study, Clinical Trial, Dataset, Meta-Analysis, Observational Study, Observational Study, Veterinary, Randomized Controlled Trial, English
\begin{figure*}[h] 
  \centering 
  \includegraphics[width=0.8\textwidth]{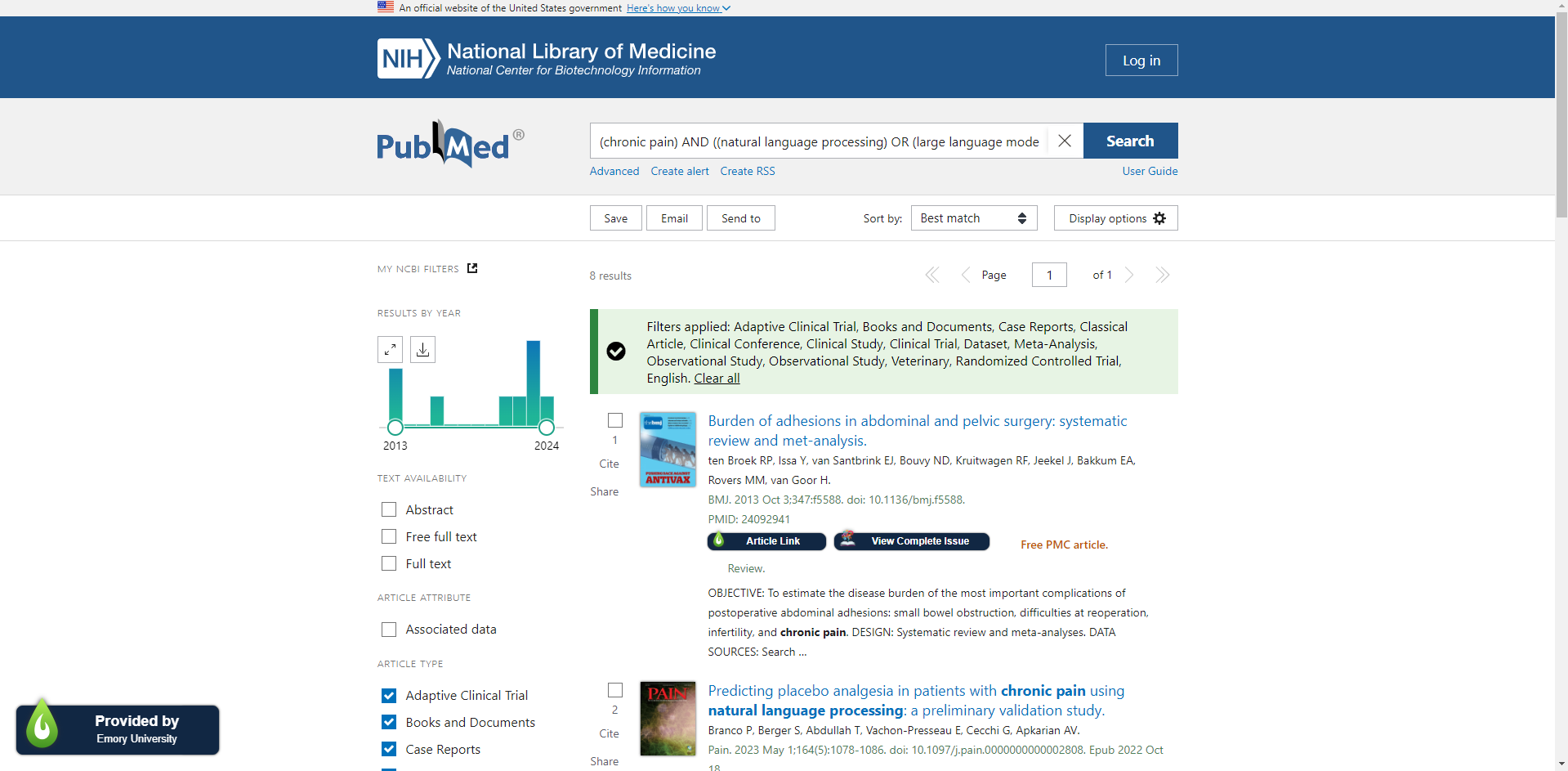}
  \caption{PubMed Search Results}
  \label{fig_pubmed}
\end{figure*}

\subsection{Web of Science}
``chronic pain" (All Fields) and ``natural language processing" OR ``large language model" (All Fields) and English (Languages)
\begin{figure*}[h] 
  \centering 
  \includegraphics[width=0.8\textwidth]{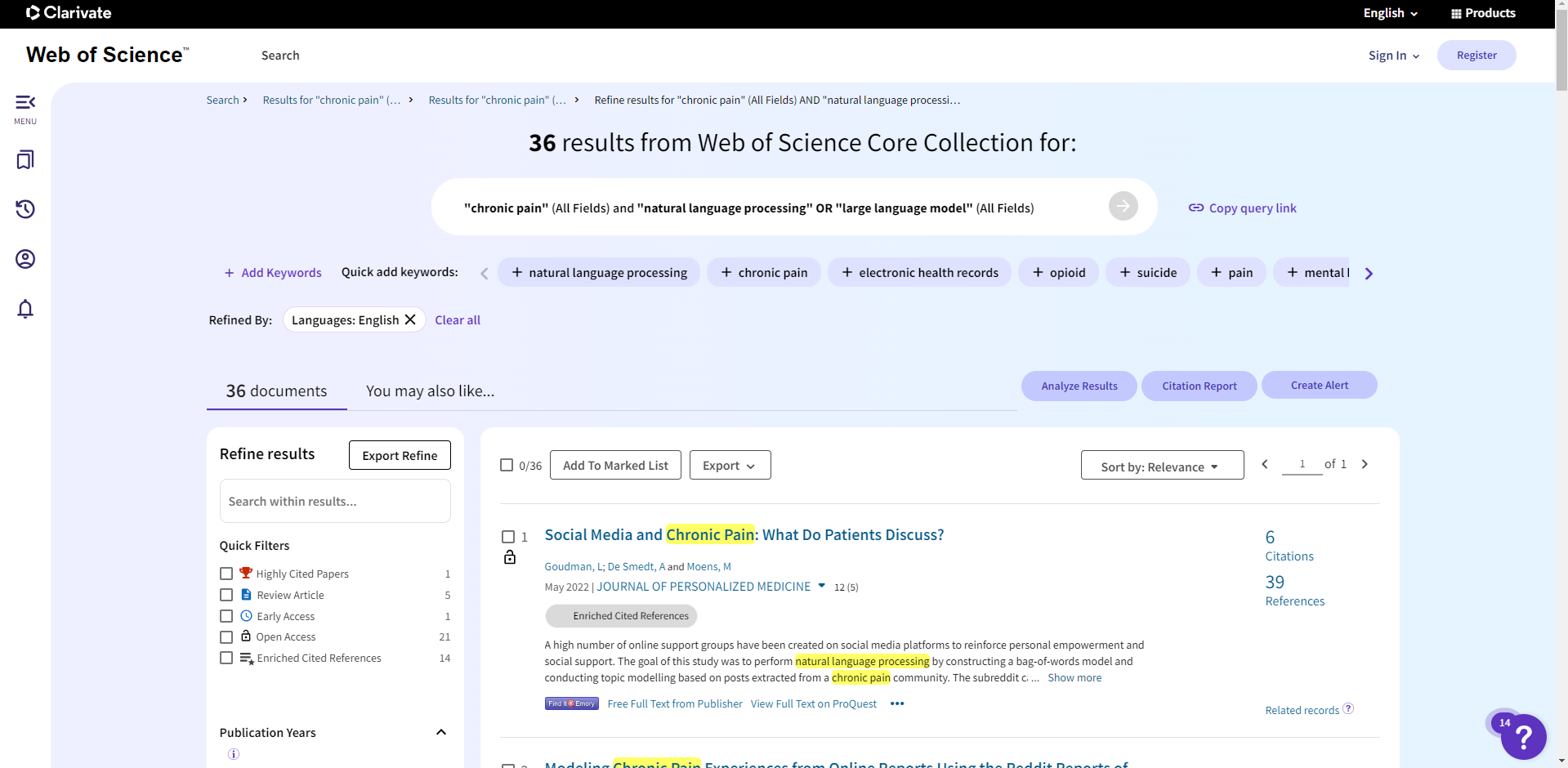}
  \caption{Web of Science Search Results}
  \label{fig_wos}
\end{figure*}

\subsection{IEEE Xplore}
(``All Metadata":``chronic pain") AND (``All Metadata":``natural language processing" OR ``All Metadata":``large language model")
 Filters Applied: 2014 - 2025
\begin{figure*}[h] 
  \centering 
  \includegraphics[width=0.8\textwidth]{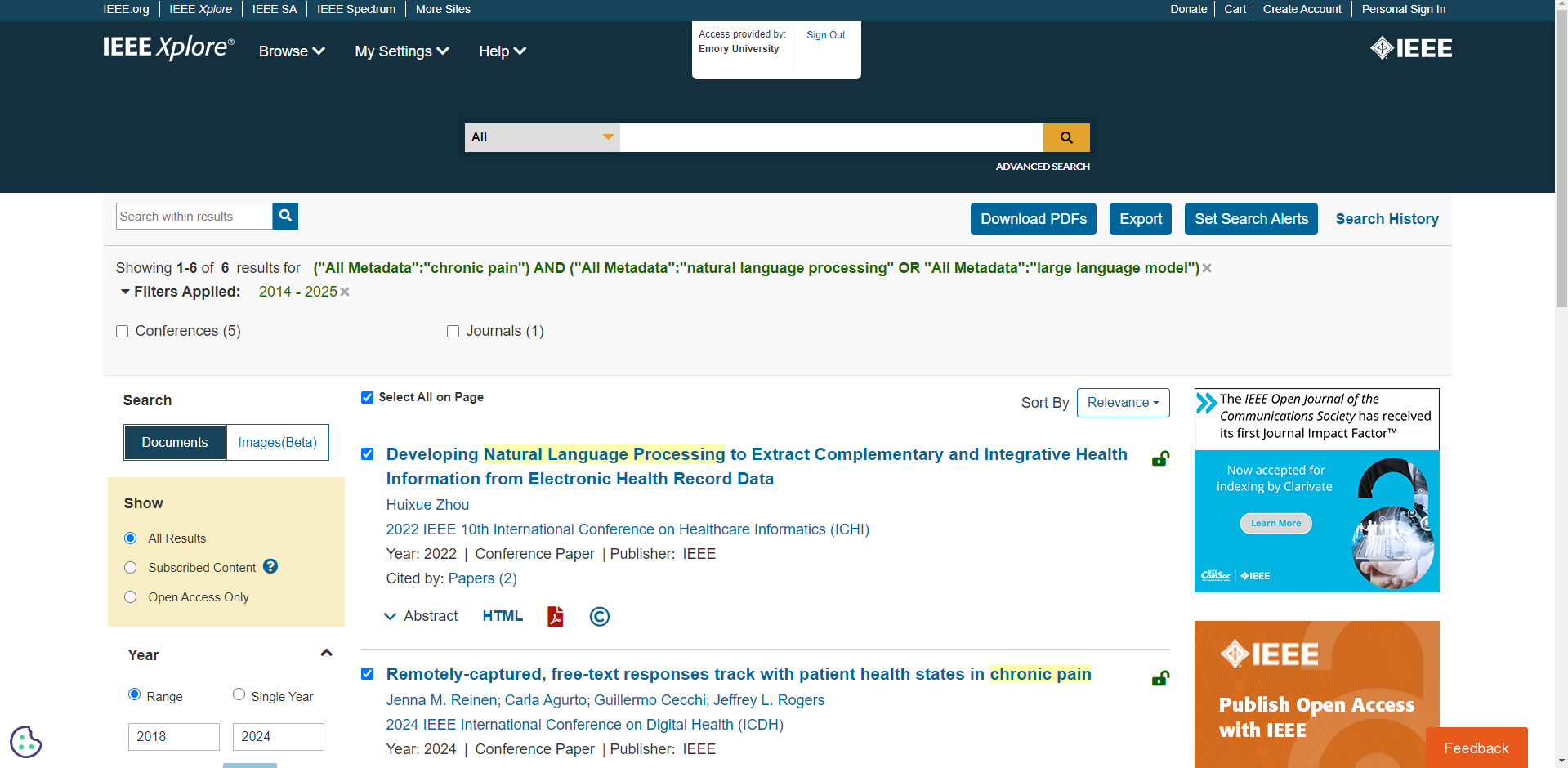}
  \caption{IEEE Xplore Search Results}
  \label{fig_ieee}
\end{figure*}

\subsection{Scopus}
Search Query:
TITLE-ABS-KEY ( ( ``chronic pain" ) AND ( ( ``natural language processing" ) OR ( ``large language model" ) ) ) AND ( LIMIT-TO ( DOCTYPE , ``ar" ) OR LIMIT-TO ( DOCTYPE , ``cp" ) ) AND ( LIMIT-TO ( LANGUAGE , ``English" ) ) AND ( LIMIT-TO ( PUBSTAGE , ``final" ) )
\begin{figure*}[h] 
  \centering 
  \includegraphics[width=0.8\textwidth]{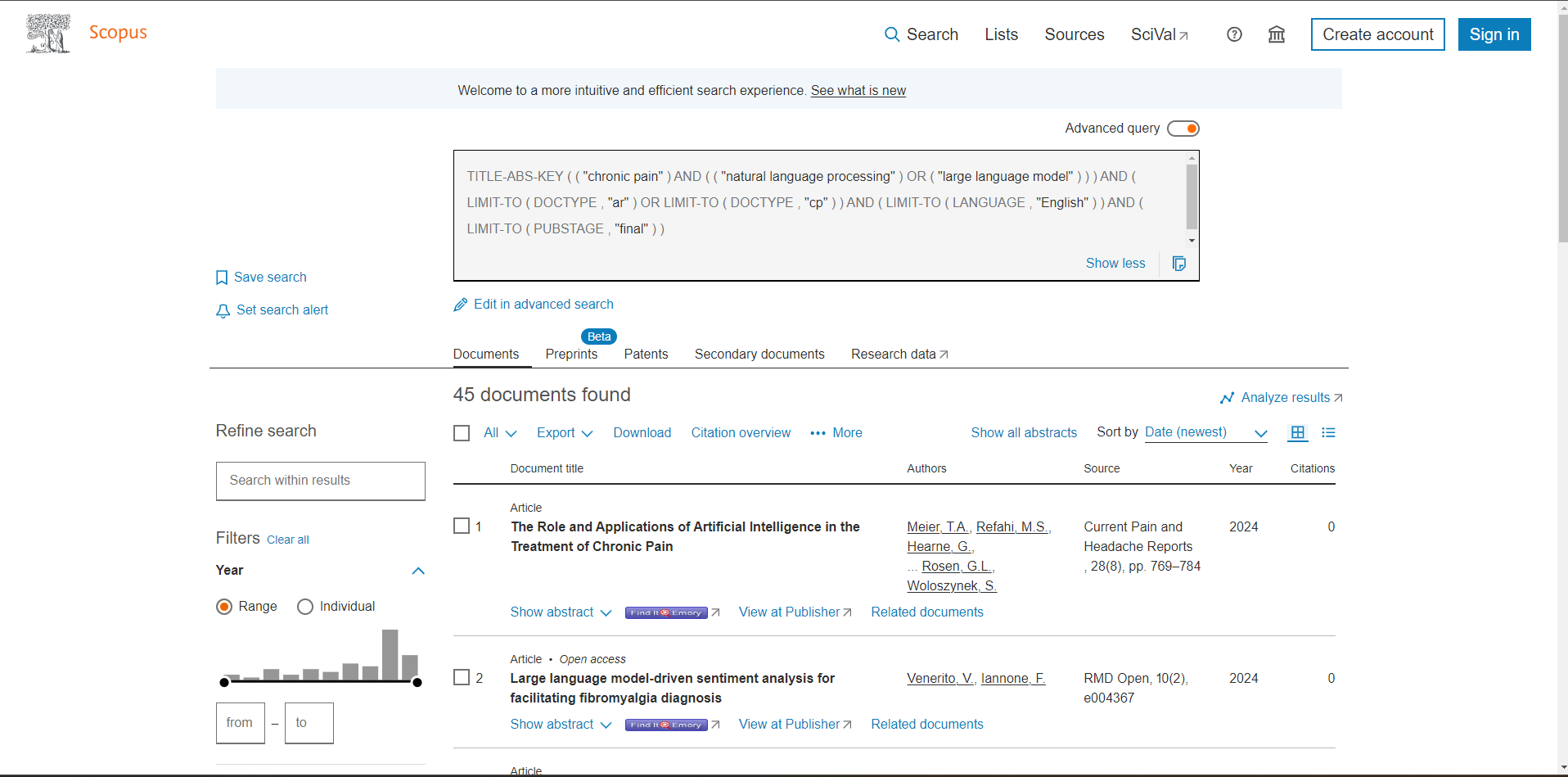}
  \caption{Scoups Search Results}
  \label{fig_scopus}
\end{figure*}

\subsection{ACL Anthology}
In ACL Anthology database, there is no way to batch export records. One has to go to ACL Anthology and enter the following search query string in the search box:
 (``chronic pain") AND (``natural language processing" OR ``large language model")
 and export the results. I used the zotero extension to export results.
 \begin{figure*}[h] 
  \centering 
  \includegraphics[width=0.8\textwidth]{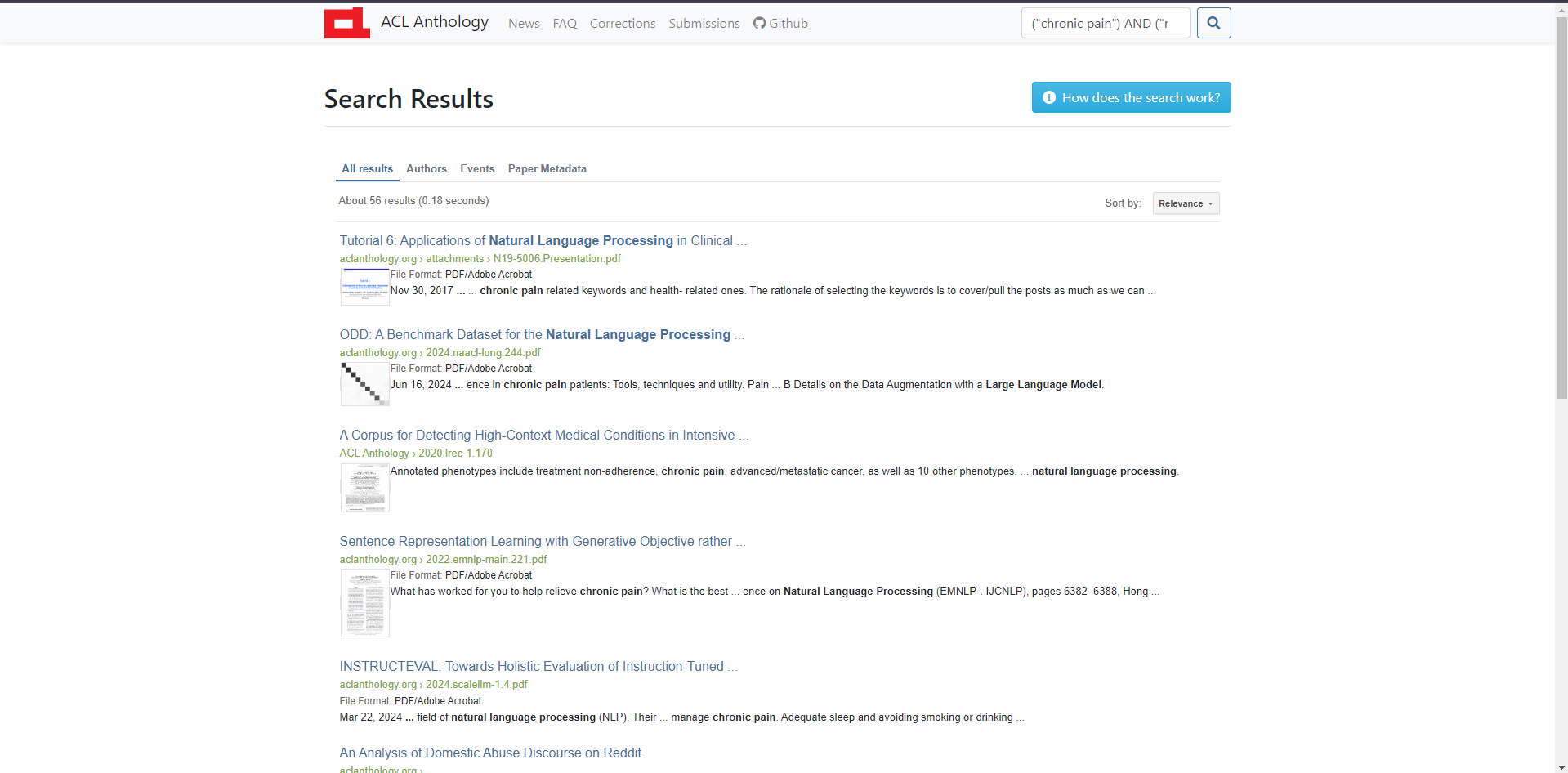}
  \caption{ACL Anthology Search Results}
  \label{fig_acl}
\end{figure*}

\section{PRISMA 2020 Checklist}
\label{checklist_table}
Please see Table~\ref{tab:checlist1} and \ref{tab:checlist2} respectively.
\begin{table}[ht]
\caption{PRISMA 2020 Checklist}
\label{tab:checlist1}
\renewcommand{\arraystretch}{1.5} 
\setlength{\tabcolsep}{4pt} 
\centering
\scriptsize
\begin{tabularx}{\textwidth}{|p{2.7cm}|p{0.6cm}|X|p{2.3cm}|}
\hline
\textbf{Section and Topic} & \textbf{Item}  & \textbf{Checklist Item} & \textbf{Item Location} \\
\hline
\multicolumn{4}{|l|}{\textbf{TITLE}} \\
\hline
Title & 1 & Identify the report as a systematic review. & \hyperref[title]{Title} \\
\hline
\multicolumn{4}{|l|}{\textbf{ABSTRACT}} \\
\hline
Abstract & 2 & See the PRISMA 2020 for Abstracts Checklist. & \hyperref[abstract]{Abstract} \\
\hline
\multicolumn{4}{|l|}{\textbf{INTRODUCTION}} \\
\hline
Rationale & 3 & Describe the rationale for the review in the context of existing knowledge. & Section~\ref{rationale} \\ 
\hline
Objectives & 4 & Provide an explicit statement of the objective(s) or question(s) the review addresses. & Section~\ref{objectives} \\
\hline
\multicolumn{4}{|l|}{\textbf{METHODS}} \\
\hline
Eligibility Criteria & 5 & Specify the inclusion and exclusion criteria for the review and how studies were grouped for the syntheses. & Section~\ref{eligibility_criteria} \\ 
\hline
Information sources & 6 & Specify all databases, registers, websites, organisations, reference lists and other sources searched or consulted to identify studies. Specify the date when each source was last searched or consulted. & Section~\ref{information_sources} \\
\hline
Search strategy & 7 & Present the full search strategies for all databases, registers and websites, including any filters and limits used. & Section~\ref{search_strategy} \\ 
\hline
Selection process & 8 & Specify the methods used to decide whether a study met the inclusion criteria, including how many reviewers screened each record, whether they worked independently, and details of automation tools used. & Section~\ref{selection_process} \\
\hline
Data collection process & 9 & Specify the methods used to collect data from reports, including how many reviewers collected data, whether they worked independently, and any processes for obtaining or confirming data from study investigators. & Section~\ref{data_collection_process} \\ 
\hline
\multirow{2}{*}{Data items} & 10a & List and define all outcomes for which data were sought. Specify whether all results that were compatible with each outcome domain in each study were sought, and if not, the methods used to decide which results to collect. & \multirow{2}{*}{Section~\ref{data_items}} \\
\cline{2-3}
 & 10b & List and define all other variables for which data were sought (e.g. participant characteristics, funding sources). Describe any assumptions made about missing or unclear information. & \\
\hline
Study risk of bias assessment & 11 & Specify the methods used to assess risk of bias in the included studies, including the tools used, how many reviewers assessed each study, and whether they worked independently. & Section~\ref{study_bias_method} \\
\hline
Effect measures & 12 & Specify for each outcome the effect measure(s) (e.g. risk ratio, mean difference) used in the synthesis or presentation of results. & N/A \\
\hline
\multirow{6}{*}{Synthesis methods} & 13a & Describe the processes used to decide which studies were eligible for each synthesis. & Section~\ref{eligibility_criteria} \\
\cline{2-4}
 & 13b & Describe any methods required to prepare the data for presentation or synthesis, such as handling of missing statistics or data conversions. & N/A \\ 
\cline{2-4}
 & 13c & Describe any methods used to tabulate or visually display results of individual studies and syntheses. & Section~\ref{data_items} \\
\cline{2-4}
 & 13d & Describe any methods used to synthesize results and provide a rationale for the choice(s). If meta-analysis was performed, describe the model(s), method(s) to identify heterogeneity, and software used. & N/A \\
\cline{2-4}
 & 13e & Describe any methods used to explore possible causes of heterogeneity among study results. & N/A \\
\cline{2-4}
 & 13f & Describe any sensitivity analyses conducted to assess robustness of the synthesized results. & N/A \\
\hline
Reporting bias assessment & 14 & Describe any methods used to assess risk of bias due to missing results in a synthesis (e.g. arising from reporting biases). & Section~\ref{bias_assessment} \\
\hline
Certainty assessment & 15 & Describe any methods used to assess certainty in the body of evidence for an outcome. & N/A \\
\hline
\end{tabularx}
\end{table}
\begin{table}[ht]
\caption{PRISMA 2020 Checklist, Cont'd.}
\renewcommand{\arraystretch}{1.5} 
\setlength{\tabcolsep}{4pt} 
\label{tab:checlist2}
\centering
\scriptsize
\begin{tabularx}{\textwidth}{|p{2.7cm}|p{0.6cm}|X|p{2.3cm}|}
\hline
\textbf{Section and Topic} & \textbf{Item}  & \textbf{Checklist Item} & \textbf{Item Location} \\
\hline
\multicolumn{4}{|l|}{\textbf{RESULTS}} \\
\hline
\multirow{2}{*}{Study selection} & 16a & Describe the results of the search and selection process, from the number of records identified in the search to the number of studies included in the review, ideally using a flow diagram. & \multirow{2}{*}{Section~\ref{study_selection}}\\
\cline{2-3}
& 16b & Cite studies that might appear to meet the inclusion criteria, but which were excluded, and explain why they were excluded. & \\
\hline
Study Characteristics & 17 & Cite each included study and present its characteristics. &  Section~\ref{study_char}\\
\hline
Risk of bias in studies & 18 & Present assessments of risk of bias for each included study. &  Section~\ref{risk_of_bias}\\
\hline
Results of individual studies & 19 & For all outcomes, present, for each study: (a) summary statistics for each group (where appropriate) and (b) an effect estimate and its precision (e.g. confidence/credible interval), ideally using structured tables or plots. &  Section~\ref{results_of_individual_studies}\\
\hline
\multirow{4}{*}{Results of syntheses} & 20a & For each synthesis, briefly summarize the characteristics and risk of bias among contributing studies. & Section \ref{study_char}\\
\cline{2-4}
& 20b & Present results of all statistical syntheses conducted. If meta-analysis was done, present for each the summary estimate and its precision (e.g. confidence/credible interval) and measures of statistical heterogeneity. If comparing groups, describe the direction of the effect. & N/A \\
\cline{2-4}
& 20c & Present results of all investigations of possible causes of heterogeneity among study results. & N/A \\
\cline{2-4}
& 20d & Present results of all sensitivity analyses conducted to assess the robustness of the synthesized results. & N/A \\
\hline
Reporting biases & 21 & Present assessments of risk of bias due to missing results (arising from reporting biases) for each synthesis assessed. & Section \ref{reprot_bias_reults}  \\
\hline
Certainty of evidence & 22 & Present assessments of certainty (or confidence) in the body of evidence for each outcome assessed. & N/A \\
\hline
\multicolumn{4}{|l|}{\textbf{DISCUSSION}} \\
\hline
\multirow{4}{*}{Discussion} & 23a & Provide a general interpretation of the results in the context of other evidence. & Section~\ref{central_findings} \\
\cline{2-4}
& 23b & Discuss any limitations of the evidence included in the review. & Section~\ref{limitation}\\
\cline{2-4}
& 23c & Discuss any limitations of the review processes used. & Section~\ref{review_limitation}\\
\cline{2-4}
& 23d & Discuss implications of the results for practice, policy, and future research. & Section~\ref{future_research}\\
\hline
\multicolumn{4}{|l|}{\textbf{OTHER INFORMATION}} \\
\hline
\multirow{3}{2.7cm}{\raggedright Registration and protocol} & 24a & Provide registration information for the review, including register name and registration number, or state that the review was not registered. & \multirow{3}{*}{N/A}\\
\cline{2-3}
& 24b & Indicate where the review protocol can be accessed, or state that a protocol was not prepared. & \\
\cline{2-3}
& 24c & Describe and explain any amendments to the information provided at registration or in the protocol. & \\
\hline
Support & 25 & Describe sources of financial or non-financial support for the review, and the role of the funders or sponsors in the review. & Section~\ref{support}\\
\hline
Competing interests & 26 & Declare any competing interests of review authors. & Section~\ref{coi_statement}\\
\hline
Availability of data, code, and other materials & 27 & Report which of the following are publicly available and where they can be found: template data collection forms; data extracted from included studies; data used for all analyses; analytic code; any other materials used in the review. & N/A\\
\hline
\end{tabularx}
\end{table}

\end{document}